\documentclass[10pt,twocolumn,letterpaper]{article}

\usepackage[pagenumbers]{cvpr} 
\usepackage{multirow}
\usepackage{caption}

\usepackage{booktabs}    
\usepackage{multirow}    
\usepackage{graphicx}    
\usepackage[table]{xcolor} 
\usepackage{amssymb}     

\definecolor{light_green}{HTML}{E6FEE6} 

\usepackage{algorithm}
\usepackage[noend]{algpseudocode}
\usepackage{amsmath}
\algnewcommand\algorithmicinput{\textbf{Input:}}
\algnewcommand\algorithmicoutput{\textbf{Output:}}
\algnewcommand\Input{\item[\algorithmicinput]}
\algnewcommand\Output{\item[\algorithmicoutput]}
\algnewcommand\Initialize{\item[\textbf{Initialize:}]}
\newcommand{\CommentLine}[1]{\State \(\triangleright\) \textit{#1}}

\usepackage{amsmath} 
\usepackage{tabularx}
\definecolor{light_green}{HTML}{C9E9C9} 

\definecolor{cvprblue}{rgb}{0.21,0.49,0.74}
\usepackage[pagebackref,breaklinks,colorlinks,allcolors=cvprblue]{hyperref}

\def\paperID{}
\def\confName{CVPR}
\def\confYear{2026}

\title{Flash-Unified: A Training-Free and Task-Aware Acceleration Framework for Native Unified Models}

\author{
Junlong Ke$^{2}$\footnotemark[1] \quad
Zichen Wen$^{1,3}$\footnotemark[1] \quad
Boxue Yang$^{1}$ \quad
Yantai Yang$^{1}$ \quad
Xuyang Liu$^{1,4}$ \quad
Chenfei Liao$^{5}$ \\
Zhaorun Chen$^{6}$\quad
Shaobo Wang$^{1}$\quad
Linfeng Zhang$^{1}$\footnotemark[2] \\
$^{1}$ Shanghai Jiao Tong University \quad
$^{2}$ Tsinghua University \quad
$^{3}$ Shanghai AI Laboratory \\
$^{4}$ Sichuan University \quad
$^{5}$ The Hong Kong University of Science and Technology (Guangzhou) \\
$^{6}$ University of Chicago \quad
}

\begin{document}
\maketitle

{
\renewcommand{\thefootnote}{\fnsymbol{footnote}}
\footnotetext[1]{Equal Contribution.}
}

{
\renewcommand{\thefootnote}{\fnsymbol{footnote}}
\footnotetext[2]{Corresponding author: zhanglinfeng@sjtu.edu.cn}
}

\begin{abstract}
Native unified multimodal models, which integrate both generative and understanding capabilities, face substantial computational overhead that hinders their real-world deployment. Existing acceleration techniques typically employ a static, monolithic strategy, ignoring the fundamental divergence in computational profiles between iterative generation tasks (e.g., image generation) and single-pass understanding tasks (e.g., VQA). In this work, we present the first systematic analysis of unified models, revealing pronounced parameter specialization, where distinct neuron sets are critical for each task. This implies that, at the parameter level, unified models have implicitly internalized separate inference pathways for generation and understanding within a single architecture. Based on these insights, we introduce a training-free and task-aware acceleration framework, \textbf{FlashU}, that tailors optimization to each task's demands.  Across both tasks, we introduce Task-Specific Network Pruning and Dynamic Layer Skipping, aiming to eliminate inter-layer and task-specific redundancy. For visual generation, we implement a time-varying control signal for the guidance scale and a temporal approximation for the diffusion head via Diffusion Head Cache. For multimodal understanding, building upon the pruned model,  we introduce Dynamic Token Pruning via a V-Norm Proxy to exploit the spatial redundancy of visual inputs. Extensive experiments on Show-o2 demonstrate that FlashU achieves 1.78$\times$ to 2.01$\times$ inference acceleration across both understanding and generation tasks while maintaining SOTA performance, outperforming competing unified models and validating our task-aware acceleration paradigm. Our code is publicly available at \url{https://github.com/Rirayh/FlashU}.
\end{abstract}

\section{Introduction}
\label{sec:intro}

With the rapid development of multi-modal large language models~\citep{Qwen2.5-VL,team2025kimi,li2024llava,wen2025efficient,wen2025ai,zhang2025docr,zhang2025trivia,team2026kimi,wen2026innovator} and image generation models~\citep{rombach2022high,labs2025flux, kang2025legion,kang2025omnidoclayout,chen2024mj}, researchers increasingly focus on integrating multimodal understanding and generation capabilities into a single model~\citep{zhang2025unified}.
The emergence of unified models has marked a paradigm shift in multimodal AI, breaking down the traditional boundaries between understanding and generative tasks within a single architecture~\citep{xie2025mme}.
However, this versatility comes at the cost of substantial computational overhead. As model parameters continue to scale, achieving efficient inference acceleration has emerged as a pivotal challenge in translating these models from research to real-world deployment~\citep{wen2025ai}.

\begin{figure}[!t]
    \centering
    \includegraphics[width=\linewidth]{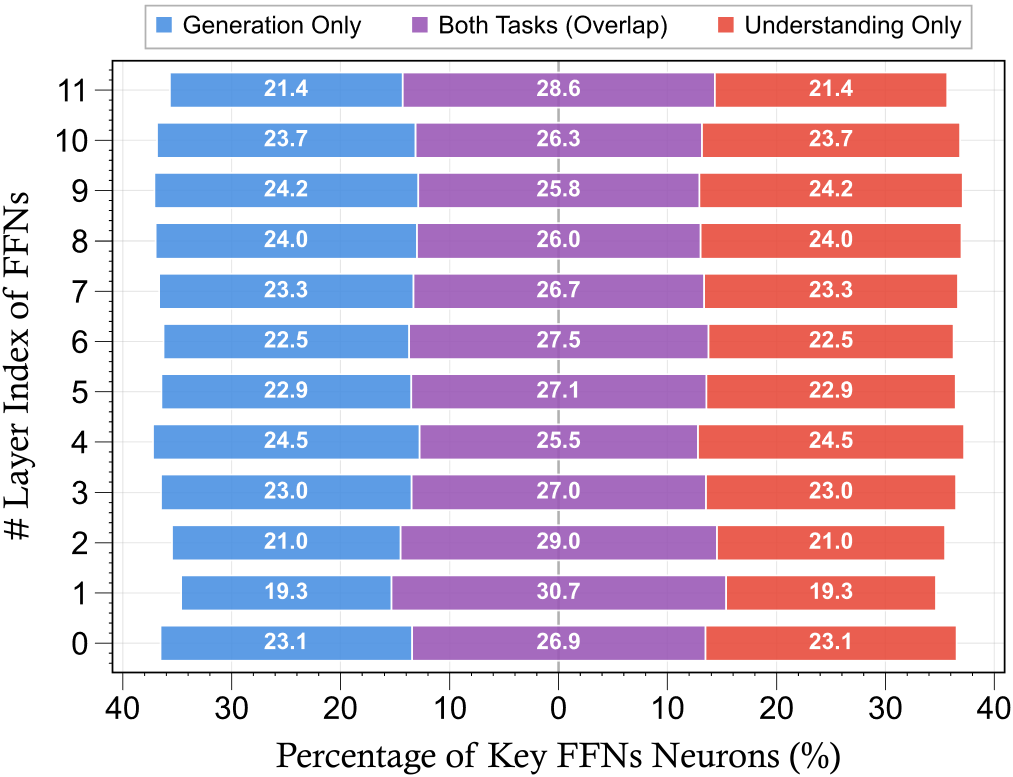}
    \vspace{-1.5em}
    \caption{
    \textbf{Task-specific FFN neuron analysis for generation and understanding.} Each layer shows the fraction of neurons that are specific to generation, understanding, or shared between both tasks (excluding unimportant neurons). Percentages indicate the proportion of neurons in each layer ranked among the top 50\% for task importance, as measured by OBD-inspired sensitivity scores $\Delta_i$.
    }
    \label{fig:ffn_neurons}
    \vspace{-1em}
\end{figure}
While significant strides have been made in model acceleration techniques~\citep{liu2025shifting,wen2025devil,sreenivas2024llm,wen2025stop,wen2025token,chen2025ipcv,he2025audiomarathon,xiong2025prune2drive,liao2025we,chen2025variation,liu2025global}, systematic investigation into redundancy analysis and efficiency optimization for unified models remain largely unexplored. Existing acceleration methods, such as structured pruning~\citep{ma2023llm} or quantization~\citep{lin2024awq}, typically adopt a \emph{static and monolithic} strategy, which rests upon an untested assumption that the model's computational requirements are homogeneous across all tasks.
We contend that this assumption overlooks a fundamental contradiction within unified models: the intrinsic computational profiles of iterative generation tasks (e.g., diffusion processes)~\citep{yang2023diffusion} and single-pass understanding tasks (e.g., VQA)~\citep{mmbench} are fundamentally distinct.
Generation tasks resemble solving ordinary differential equations (ODEs)~\citep{lu2022dpm}, representing a multi-step denoising process that evolves from blurry to clear, where early steps exhibit high fault tolerance. Conversely, understanding tasks rely on the hierarchical abstraction and evolution of features within a single forward pass. 
Imposing a uniform acceleration strategy on unified models inevitably leads to a trade-off between competing objectives.

\begin{figure}[!h]
\vspace{-0.7em}
    \centering
    \includegraphics[width=\linewidth]{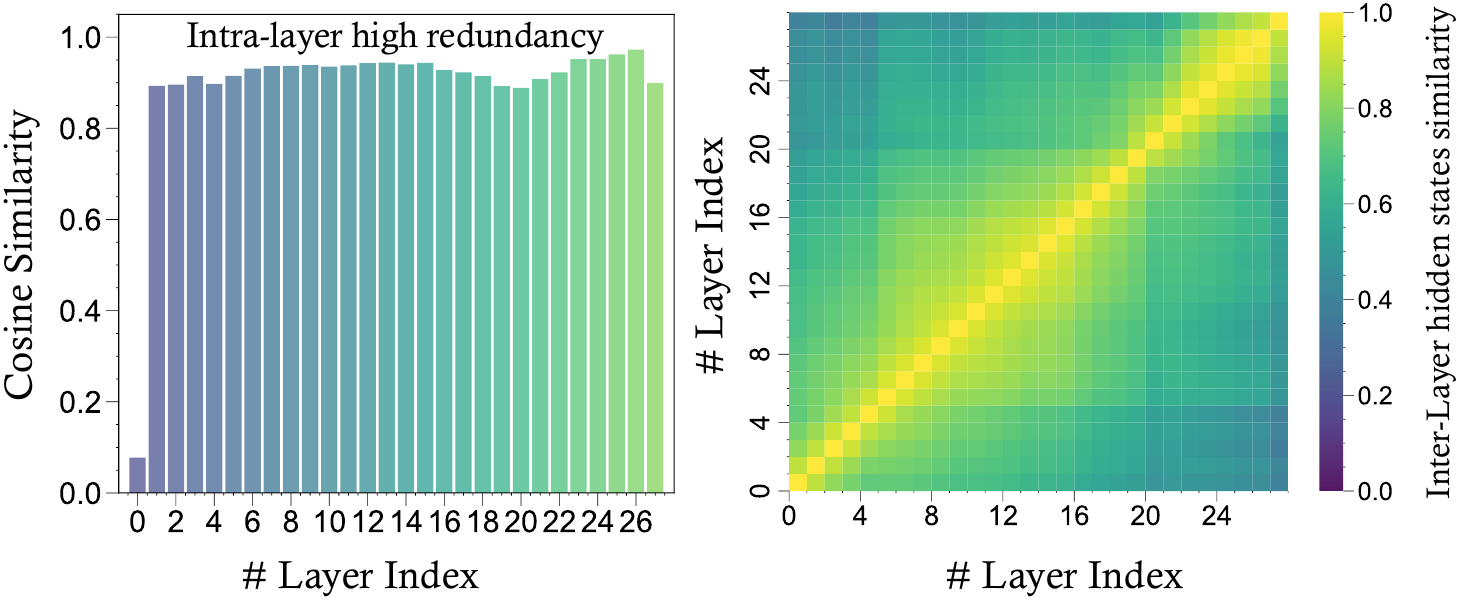}
\vspace{-1.5em}
\caption{\textbf{Understanding Redundancy.} \textbf{Left:} Intra-layer cosine similarity between the input and output features of each layer for newly generated tokens in understanding tasks. \textbf{Right:} Inter-layer heatmap of the cosine similarity among layer outputs for newly generated tokens. All measurements were conducted on samples from the MME~\cite{mme}.}
    \label{fig:under_redundancy}
    \vspace{-0.8em}
\end{figure}
To bridge this gap, we present the first systematic dissection of the internal computational mechanisms of unified models, revealing two pivotal phenomena. First, we observe pronounced parameter specialization: the model's feed-forward networks (FFNs) contain numerous neurons exclusively important for either generation or understanding tasks, with only a limited proportion of shared critical neurons (Figure~\ref{fig:ffn_neurons}). This indicates that despite a unified architecture, the model spontaneously develops task-specific computational pathways internally.
Second, inter-layer similarity analysis reveals that generation tasks exhibit exceptionally high feature redundancy across layers during the denoising process, while key tokens in understanding tasks demonstrate progressively evolving features with increasing depth, rendering them more sensitive to intermediate layer pruning (Figure~\ref{fig:under_redundancy}).

These findings compellingly demonstrate that for unified models, an ``optimal'' acceleration strategy must be task-aware and dynamic. 

Building upon this insight, we introduce \textbf{FlashU}, a training-free and task-aware acceleration framework for unified generative-understanding models. Our core principle is that acceleration strategies should align with, rather than oppose, the model's internal parameter specialization mechanisms. For generation tasks, we conceptualize the process as a \emph{coarse-to-fine} optimization problem, designing a Hybrid FFN that dynamically switches between lightweight and full paths during different generation phases, coupled with an adaptive ODE solver featuring multi-stage guidance and iterative caching. For understanding tasks, we combine static pruning of generation-exclusive neurons with dynamic visual token pruning to eliminate spatial redundancy. The distinguishing characteristic of our framework is its orthogonality: static pruning in understanding tasks preserves generative capacity while coexisting seamlessly with dynamic pathways in generation tasks through shared-weight design.
In summary, our contributions are threefold:
\begin{itemize}
    \item We conduct the first systematic investigation of internal computational redundancy in unified generative-understanding models, revealing the phenomena of parameter specialization and computational heterogeneity.
    \item We introduce the first training-free and task-aware acceleration framework for native unified models, enabling adaptive acceleration that aligns with the model's internal inference pathways.
    \item Extensive experiments on Show-o2 demonstrate that our framework achieves 1.78$\times$ to 2.01$\times$ inference acceleration across both understanding and generation tasks while maintaining SOTA performance, outperforming competing unified models.
\end{itemize}

\begin{figure*}[t]
    \centering
    \includegraphics[width=\linewidth, trim=1cm 7.8cm 7.9cm 3.5cm, clip]{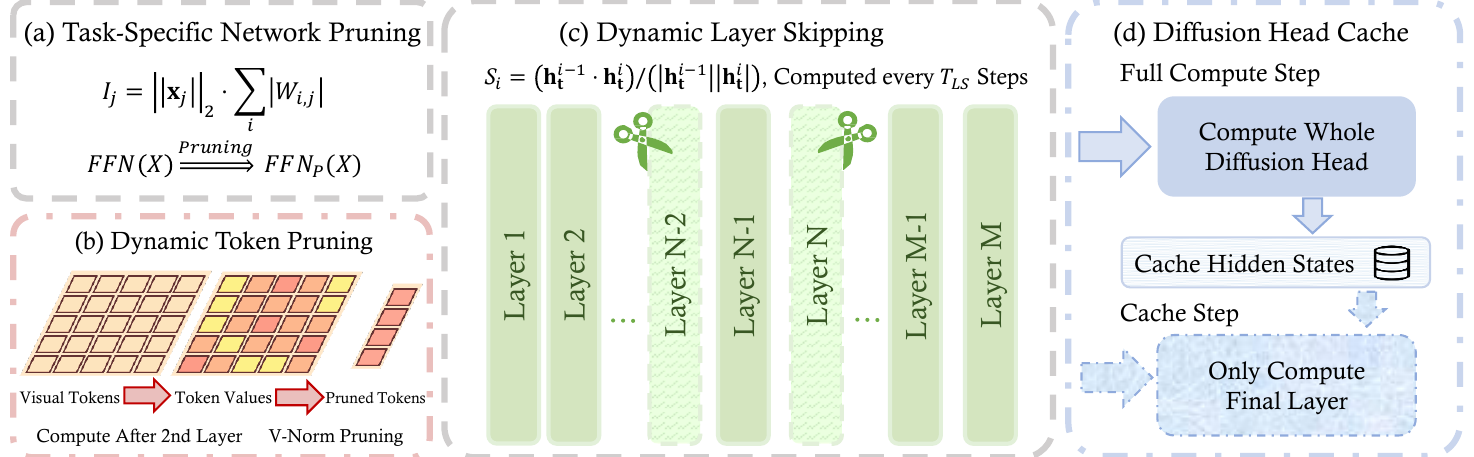}

    \caption{\textbf{Illustration of the FlashU acceleration framework.} 
    (a) \textbf{Task-Specific FFN Pruning} calculates an importance score $I_j$ based on activation norms and weight magnitude to statically mask redundant neurons for a specific task. 
    (b) \textbf{Dynamic Token Pruning} utilizes the V-Norm proxy at shallow layers (the 2nd layer) to identify and prune visually spatially redundant tokens. 
    (c) \textbf{Dynamic Layer Skipping} bypasses layers exhibiting high input-output cosine similarity ($S_i$), re-evaluating the skipping mask every $T_{LS}$ steps. 
    (d) \textbf{Diffusion Head Cache} exploits temporal coherence in the generative process, computing the full diffusion head only at fixed intervals and reusing cached hidden states for intermediate steps.}
    \label{fig:flashu_modules}
\end{figure*}

\section{Related Work}
\label{sec:related_works}

\noindent \textbf{Multimodal Large Language Models.}
Building on the significant progress of Large Language Models (LLMs)~\cite{llama, qwen2.5}, Multimodal Large Language Models (MLLMs)~\cite{llava, internvl, Qwen2.5-VL} have demonstrated remarkable capabilities in general-purpose visual understanding and reasoning. The predominant architecture utilizes a pre-trained vision encoder to extract visual features, which are subsequently projected and aligned with the LLM's embedding space. Concurrently, a subset of MLLMs, known as encoder-free models~\cite{showo, diao2024EVE, diao2025EVEv2}, has emerged, aiming to directly align raw visual features within this space. However, these methods typically trail in performance compared to their counterparts that use aligned image-text features. Beyond architectural considerations, recent research~\cite{sharegpt4v, tong2024cambrian1, li2024llava} has emphasized the critical role of high-quality, large-scale instruction-tuning data in advancing MLLM capabilities.

\noindent \textbf{Unified Multimodal Models.}
Unified Multimodal Models (UMMs) represent an emerging research area focused on integrating the distinct capabilities of Large Multimodal Models (LMMs) and visual generative models into a single, cohesive system. This field has primarily advanced along two paradigms. The first approach involves developing native unified models~\cite{team2024chameleon, showo, zhou2025transfusion, wang2024emu3, vila-u}, which feature a unified architecture inherently designed for both multimodal understanding and generation, often employing autoregressive modeling, diffusion, or a hybrid of these techniques. Subsequent research in this area has focused on refining training pipelines and enhancing the semantic richness of discrete visual tokens~\cite{unitok, unitoken, SemHiTok, dualtoken}. The second paradigm involves assembling specialized, off-the-shelf components~\cite{CoDI, uio2, dreamllm, seed-x, tong2024metamorph, metaqueries, blip3}. This method connects pre-trained LMMs for understanding with visual generative models for generation, typically by fine-tuning lightweight adapters or learnable query tokens. Representative works~\cite{seed-x, metaqueries, blip3} have demonstrated the significant capabilities of these assembled frameworks, underscoring their potential as an efficient pathway toward unification.

\section{Methodology}
\label{sec:method}

\subsection{Overview}
\label{subsec:overview}

Our approach is founded on the principle of aligning optimization with the model's intrinsic parameter specialization, wherein distinct neural circuits are dedicated to either generation or understanding tasks.

We first implement network redundancy elimination. A unified Feed-Forward Network (FFN) pruning policy accelerates both generation and understanding, while retaining the original FFNs to form a Hybrid FFN. 
Furthermore, we introduce Dynamic Layer Skipping. This strategy specifically targets and eliminates inter-layer redundancy, thereby augmenting the redundancy elimination achieved by the preceding FFN method.

Beyond these foundational redundancy elimination, FlashU establishes two orthogonal, non-interfering acceleration paths within the single model, rather than imposing a uniform optimization. This dual-mode architecture allows the model to dynamically select the most efficient computational graph based on the task's unique demands.

For iterative generative tasks, the framework employs an adaptive ODE solver augmented with dynamic guidance scaling and diffusion head iterative caching.

For single-pass understanding tasks, it utilizes visual token pruning in addition to the aforementioned network redundancy elimination techniques, which effectively mitigates spatial and computational redundancy.


\subsection{Motivation}

\subsubsection{Parameter Specialization}
\label{subsec:analysis}

Our investigation begins with a fundamental question: \textbf{\emph{do unified models utilize their parameters monolithically across different tasks?}} Traditional acceleration methods implicitly assume so, applying a uniform optimization strategy. We challenge this assumption by analyzing the phenomenon of parameter specialization. 

To quantify the task-specific importance of individual neurons, we employ a sensitivity analysis method inspired by Optimal Brain Damage (OBD). While traditional OBD assesses parameter saliency based on the global loss, we adapt this concept to evaluate a neuron's contribution to the layer's local output. For each neuron \(i\) in an FFN, we compute an importance score \(\Delta_i\) as the reconstruction error incurred by zeroing out that neuron's activation: 
\begin{equation}
\label{eq:obd_score}
\Delta_i = \mathbb{E}_{x \sim \mathcal{D}} \left[ \left\| \mathrm{FFN}(x) - \mathrm{FFN}_{-i}(x) \right\|_2^2 \right]
\end{equation}

where \(\mathcal{D}\) is a calibration dataset, \(\mathrm{FFN}(x)\) is the layer's original output, and \(\mathrm{FFN}_{-i}(x)\) is its output with neuron \(i\) zeroed out. This directly measures the expected local feature degradation if that neuron were to be removed. We compute two separate scores for each neuron: a generation score \(\Delta_i^G\) using a subset of the text-to-image generation dataset (DPG-Bench~\cite{dpgbench}), and an understanding score \(\Delta_i^U\) using a subset of the VQA dataset (MME~\cite{mme}).

Our analysis reveals a pronounced pattern of parameter specialization within the FFNs in both generation and understanding tasks. We observe that a significant portion of neurons are exclusively critical for either generation (high \(\Delta_i^G\), low \(\Delta_i^U\)) or understanding (low \(\Delta_i^G\), high \(\Delta_i^U\)). As shown in Figure~\ref{fig:ffn_neurons}, only a small subset of neurons are important for both tasks, confirming that the model forms distinct pathways internally. This insight motivates our framework: effective acceleration should leverage such specialization by designing dedicated optimization paths aligned with task-specific structures.
\subsubsection{Temporal Redundancy}

\begin{figure}[!h]
\vspace{-1em}
    \centering
    \includegraphics[width=\linewidth]{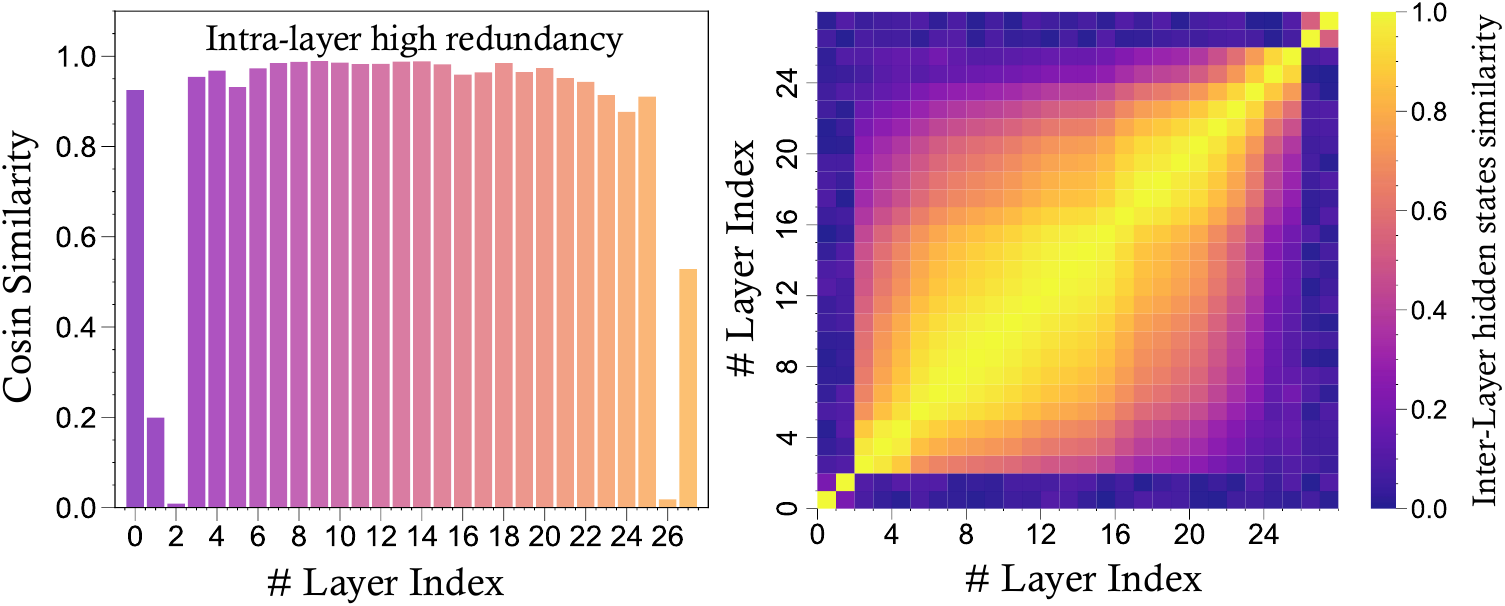}
    \vspace{-1.5em}

\caption{\textbf{Redundancy in Generation Tasks.} \textbf{Left:} Intra-layer cosine similarity between the input and output of each layer. \textbf{Right:} Inter-layer heatmap of hidden state cosine similarity. All measurements were conducted on samples from the  GenEval~\cite{geneval}.}
    \label{fig:gen_redundancy}
    \vspace{-1.5em}
\end{figure}

While the unified model operates in distinct modes (full self-attention for generation and causal self-attention for understanding), our analysis reveals a common opportunity for optimization. As illustrated in Figure~\ref{fig:gen_redundancy} and Figure~\ref{fig:under_redundancy}, we observe great inter-layer redundancy in both tasks by visualizing the similarity between layer inputs and outputs. 

\subsection{Task-Specific Network Pruning}
Our network pruning methodology is bifurcated into two complementary strategies: Task-Specific FFN Pruning and Dynamic Layer Skipping. The former addresses intra-layer redundancy by applying a one-shot, task-aware pruning mask to FFNs, while preserving the vanilla weights for hybrid computation. The latter addresses inter-layer redundancy by dynamically bypassing layers based on an input-output similarity metric, which is re-evaluated at intervals.

\subsubsection{Task-Specific FFN Pruning}
Our design for generation acceleration is motivated by the observation that parameters within the unified model exhibit a high degree of task specialization. As illustrated in Figure~\ref{fig:ffn_neurons}, distinct sets of neurons are critical exclusively for either generation or understanding tasks. This specialization implies that when executing a given task, the full computational power of the FFN layers is not required, as a substantial portion of the parameters may be non-essential. 
To balance computational cost and quality, we introduce a \textbf{Hybrid FFN} architecture. This architecture consists of both a \textit{full path} ($\text{FFN}_{\text{f}}$) and a \textit{pruned path} ($\text{FFN}_{\text{p}}$), allowing the model to alternate between them based on the timestep $t$.



Inspired by Wanda~\cite{sun2023simple}, we compute an importance score $I_j$ for each neuron $j$. This score is the product of the component's aggregate weight magnitude and its corresponding mean input activation norm:
\begin{equation}
\label{eq:wanda}
I_j = ||\mathbf{x}_{j}||_2 \cdot \sum_i |W_{i,j}|.
\end{equation}
The activation norms $||\mathbf{x}_j||_2$ are gathered empirically: For image generation tasks, we collect these statistics during the generation process of the first image; For understanding tasks, we compute the norms at initialization using a 20-sample subset of the MME dataset.

This score $I_j$ is used to create a static mask $\mathbf{M}$ that defines the pruned path $\text{FFN}_{\text{p}}$. We prune the $r_p$ proportion of FFN hidden dimension parameters, corresponding to the neurons with the lowest $r_p$ ratio of $I_j$ scores. The operation is defined as:
\begin{equation}
\label{eq:pruned_ffn_def}
\text{FFN}_{\text{p}}(x) = W_2 (\mathbf{M} \odot \sigma(W_1 x + b_1)) + b_2.
\end{equation}
where $\sigma$ is the activation function and $\odot$ denotes element-wise multiplication. The final output of the Hybrid FFN is formulated using an indicator function $\mathbf{1}(\cdot)$ based on a switching threshold $\tau$ (defaulting to 0.2):
\begin{equation}
\label{eq:hybrid_ffn_elegant}
y_{\text{h}}(x, t) = \mathbf{1}(t \le \tau T) \cdot \text{FFN}_{\text{f}}(x) + \mathbf{1}(t > \tau T) \cdot \text{FFN}_{\text{p}}(x).
\end{equation}
This significantly reduces the expected computational cost $\mathbb{E}[C]$ over the total duration $T$:
\begin{equation}
\label{eq:expected_cost}
\mathbb{E}[C] = \tau C_{\text{f}} + (1-\tau) C_{\text{p}}.
\end{equation}

\subsubsection{Dynamic Layer Skipping}
To address the inter-layer redundancy illustrated in Figure~\ref{fig:gen_redundancy} and Figure~\ref{fig:under_redundancy}, we designed the \textbf{Dynamic Layer Skipping} strategy, which dynamically identifies and bypasses redundant layers within fixed intervals.
A layer's contribution is quantified by measuring its functional transformation. We define a similarity score $S_i$ for each layer $i$, computed via cosine similarity between its input and output.
For \textbf{generation tasks}, which update the entire sequence via full self-attention, we compute the mean cosine similarity between a layer's input hidden state $\mathbf{H}_i$ and its output hidden state $\mathbf{H}_{i+1}$ across a batch $D$:
\begin{equation}
S_i^{\text{gen}} = \frac{1}{D} \sum_{d=1}^{D} \frac{\mathbf{H}_{i}^{(d)} \cdot \mathbf{H}_{i+1}^{(d)}}{ \|\mathbf{H}_{i}^{(d)}\|_2 \cdot \|\mathbf{H}_{i+1}^{(d)}\|_2 }
\end{equation}
A similarity score $S_i^{\text{gen}}$ approaching $1$ signifies high redundancy, as the layer's output vector is directionally similar to its input, implying a minimal transformation.
Conversely, for \textbf{understanding tasks}, which employ causal self-attention, computation is focused on the newly generated token. We therefore measure similarity by considering only the transformation of the hidden state at the final token position, $\mathbf{h}_{t_L}$. The score is computed across the batch $D$:
\begin{equation}
S_i^{\text{und}} = \frac{1}{D} \sum_{d=1}^{D} \frac{\mathbf{h}_{i, t_L}^{(d)} \cdot \mathbf{h}_{i+1, t_L}^{(d)}}{ \|\mathbf{h}_{i, t_L}^{(d)}\|_2 \cdot \|\mathbf{h}_{i+1, t_L}^{(d)}\|_2 }
\end{equation}
The skipping policy is applied dynamically and updated periodically. We define a recalculation interval, $T_{LS}$. In the full-computation step of each interval, the model executes a forward pass using all layers. This step is leveraged to re-compute the similarity score $S_i$ for every layer, leading to the generation of a new skip list, $\mathcal{L}_{\text{skip}}$. This list contains the indices of the layers with the highest similarity scores (i.e., the most redundant layers) determined by a predefined skipping ratio $r_{LS}$. The derived skip list $\mathcal{L}_{\text{skip}}$ is cached and subsequently applied for the duration of the following $T_{LS}-1$ steps, enabling the model's forward pass to bypass the full computation for any layer $i \in \mathcal{L}_{\text{skip}}$.

\subsection{Generation Acceleration}
\label{subsec:generation_accel}

The acceleration for iterative denoising generation tasks is designed around a coarse-to-fine optimization principle, viewing the process as the numerical solution to an Ordinary Differential Equation (ODE). Our approach introduces two synergistic optimizations that enhance this numerical solution: time-varying control signal for guidance scale and a temporal approximation for diffusion head via diffusion head cache.

\subsubsection{Adaptive Guidance Scale}

Standard approaches employ a static, high guidance scale $s$, a known trade-off that often introduces over-saturation or requires a high Number of Function Evaluations (NFE) to converge. The design of $s(t)$ is critical and non-trivial. Recent work~\cite{jin2025stage} has established that the diffusion process is stage-wise, beginning with a ``Mode Selection'' phase (high $t$, high noise) and concluding with a ``Concentration'' phase (low $t$, low noise). A key finding is that strong, early guidance can be detrimental, as it ``erodes global diversity''~\cite{jin2025stage} by forcing a premature collapse during the Mode Selection phase.
Based on this understanding, we propose a principled, ascending guidance schedule $s(t)$ that begins with a low-gain regime and transitions to a high-gain regime:
\begin{equation}
\label{eq:guidance_schedule_v}
s(t) = s_{\text{low}} \cdot \mathbf{1}(t > t_{\text{switch}}) + s_{\text{high}} \cdot \mathbf{1}(t \le t_{\text{switch}}).
\end{equation}
This strategy is strongly supported. Our use of low-gain control ($s_{\text{low}}$) during the early, high-noise stage ($t > t_{\text{switch}}$) respects the stochastic nature of the Mode Selection phase, preserving global diversity~\cite{jin2025stage} while preventing the ``overshooting'' artifacts associated with static guidance~\cite{wang2024analysis}. The subsequent switch to high-gain control ($s_{\text{high}}$) during the late-stage ($t \le t_{\text{switch}}$) effectively amplifies within-mode contraction, aligning with the ``Concentration'' phase to refine fidelity.
This dynamic, monotonically increasing schedule, which~\cite{wang2024analysis} empirically demonstrates improves performance—is not just a qualitative improvement; it is a key enabler for acceleration. By increasing the guidance efficacy in the final steps, it allows the model to achieve high fidelity with significantly fewer total steps. Furthermore, this strategy creates a powerful computational synergy. 

\subsubsection{Diffusion Head Cache}
The Diffusion Head component, responsible for predicting the velocity $\hat{v}_\theta$ of the denoising flow, often contains a series of computationally intensive layers. To mitigate this overhead, we introduce the Diffusion Head Cache, a mechanism designed to leverage the inherent temporal coherence observed across successive diffusion steps by periodically reusing expensive computations.

The flow dynamically selects one of two operational pathways at each timestep $t$ based on a predefined cache interval, $\mathcal{T}_{cache}$. When the current step number is a multiple of the cache interval, the model executes the standard, full computation and cache the hidden state before the last layer. 
In all other subsequent steps (i.e., $t \pmod{\mathcal{T}_{cache}} \neq 0$), the model bypasses the costly Diffusion Head layers entirely. Instead, the required processed hidden state is read directly from the cache, effectively utilizing the result computed during the last full-computation step. This temporal approximation significantly reduces inference latency. 





\subsection{Understanding Acceleration}
\label{subsec:understanding_accel}

\begin{figure}[!t]
    \centering
    \includegraphics[width=\linewidth]{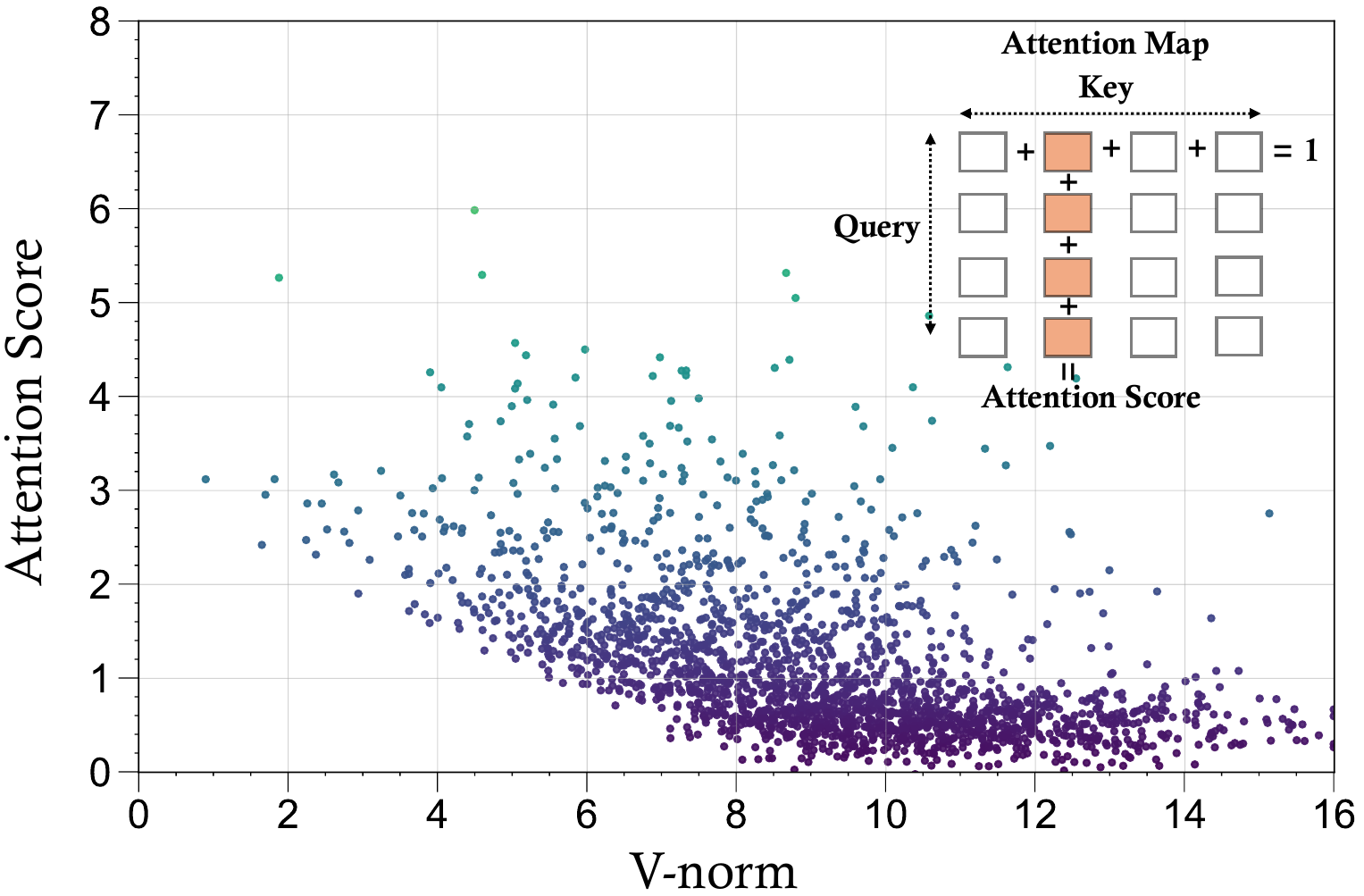}
    \vspace{-1.6em}
    \caption{The relation between attention score and the norm of the value matrix for each token. Tokens with the largest attention score tend to have a lower norm in the value matrix.}
   \label{fig:attention_score_vs_v-norm}
    \vspace{-1.6em}
\end{figure}

\subsubsection{Dynamic Token Pruning via V-Norm Proxy}

Building upon the pruned model, we introduce a dynamic mechanism to exploit the spatial redundancy of visual inputs. While prevailing token pruning methods rely on attention scores~\citep{chen2024image}, this approach is incompatible with modern, optimized implementations like Flash Attention~\citep{dao2022flashattention,dao2023flashattention2}, which avoid materializing the full attention matrix. Our solution is an efficient proxy for token importance. We empirically observe a consistent negative correlation between a token's attention score and the L2 norm of its value vector (v-norm), as shown in Figure~\ref{fig:attention_score_vs_v-norm}. This indicates that tokens with high attention scores (i.e., high importance) tend to have smaller v-norms.

Leveraging this insight, we use the v-norm as a low-cost, Flash Attention-compatible proxy for visual token pruning. In a shallow layer (e.g., the second layer), we compute the importance score \(I_j\) for each visual token \(\mathbf{t}_j\) as the L2 norm of its value vector \(\mathbf{v}_j\): $I_j = \| \mathbf{v}_j \|_2$.
We then prune a predefined ratio \(\rho\) of the tokens with the highest v-norm scores. 

\begin{table*}[!h]

    \centering
    \caption{Evaluation on multimodal understanding benchmarks. \# Params. indicates the number of parameters of the base LLM. Results in \textcolor{gray}{gray} indicate the vanilla Show-o2 performance. Latency is measured in seconds (s), conducted on samples from MMMU.}
        \vspace{-0.5em}

    \label{tab:mmu_comparison}
    \renewcommand{\arraystretch}{0.93}
    \setlength\tabcolsep{10pt}
    \resizebox{\linewidth}{!}{%
    \begin{tabular}{lccccccc|c}
            \toprule
            \multirow{2}{*}{Models} & \multirow{2}{*}{\# Params.} & MME $\uparrow$ & \multirow{2}{*}{GQA$ \uparrow$} & MMB $\uparrow$ & MMMU $\uparrow$ & \multirow{2}{*}{MMStar $\uparrow$} & \multirow{2}{*}{AI2D $\uparrow$} & \multirow{2}{*}{\textbf{Latency (s)} $\downarrow$}\\
            & & (p) & & (en) & (val) & & \\ 
            \midrule
     MetaMorph~\cite{tong2024metamorph} 			& 8B 						&- 									& - 					& 75.2 						& - 						& - 						& -  & - \\
     ILLUME~\cite{ILLUME} 							& 7B 						& 1445.3 							& - 					& 75.1 						& 38.2 						& - 						& 71.4 & - \\
     Show-o~\cite{showo} 							& 1.3B 						& 1097.2 							& 58.0 					& - 						& 27.4 						& - 						& - & -\\
     JanusFlow~\cite{ma2024janusflow} 				& 1.5B 						& 1333.1 							& 60.3 					& 74.9 						& 29.3 						& - 						& - & -\\
     SynerGen-VL~\cite{synergen-vl} 				& 2.4B 						& 1381.0 							& - 					& 53.7 						& 34.2 						& - 						& - & -\\
     Liquid~\cite{liquid} 							& 8B 						& 1448.0 							& 61.1 					& - 						& - 						& - 						& - & -\\
    
            Emu3~\cite{wang2024emu3} & 8B & - & 60.3 & 58.5 & 31.6 & - & 70.0 & 1.29\\ 
            VILA-U~\cite{vila-u} & 7B & 1317.4 & 58.5 & 63.5 & 30.7 & 37.2 & 48.9 & 0.12\\ 
            \midrule
            Show-o2~\cite{xie2025show} & 1.5B & \textcolor{gray}{1450.9} & \textcolor{gray}{60.0} & \textcolor{gray}{67.4} & \textcolor{gray}{37.1} & \textcolor{gray}{43.4} & \textcolor{gray}{69.0} & \textcolor{gray}{-} \\ 
            Show-o2~\cite{xie2025show} & 7B & \textcolor{gray}{1620.5} & \textcolor{gray}{63.1} & \textcolor{gray}{79.3} & \textcolor{gray}{48.9} & \textcolor{gray}{56.6} & \textcolor{gray}{78.6} & \textcolor{gray}{1.71} \\ 
            \rowcolor{green!10}
            \hspace{0.6em} + \textbf{FlashU (Ours)} & 7B & 1560.5 & 60.4 & 75.3 & 45.1 & 48.3 & 73.3 & 0.96\textsubscript{+1.78$\times$} \\ 
    
            \bottomrule
    \end{tabular}
    }
    \vspace{-1em}
    \end{table*}

\section{Experiments}
\label{sec:exp}

\begin{table*}[!ht]
    \centering
    \caption{Evaluation on the GenEval~\cite{geneval}. \# Params. indicates the number of parameters of the base LLM. Results in \textcolor{gray}{gray} indicate the vanilla Show-o2 performance. The subscript values show the speedup achieved by FlashU. Latency is measured in seconds (s).}
    \vspace{-0.5em}
    \label{tab:geneval_benchmark}
    \renewcommand{\arraystretch}{0.95}
    \resizebox{\linewidth}{!}{%
    \begin{tabular}{lcccccccc|c}
            \toprule
          Method & \# Params. & Single Obj. & Two Obj. & Counting  & Colors &  Position & Color Attri. & \textbf{Overall}$\uparrow$ & \textbf{Latency (s)}$\downarrow$ \\
          \midrule
     SEED-X~\cite{seed-x} 					& 17B 						& 0.97 				& 0.58 				& 0.26 				& 0.80 				& 0.19 				& 0.14 				& 0.49 & -\\
     TokenFlow-XL~\cite{qu2024tokenflow} & 14B 				& 0.95 				& 0.60 				& 0.41 				& 0.81 				& 0.16 				& 0.24 				& 0.55 & -\\
     ILLUME~\cite{ILLUME} 					& 7B 					& 0.99 				& 0.86 				& 0.45 				& 0.71 				& 0.39 				& 0.28 				& 0.61 & -\\
     Show-o~\cite{showo} 				& 1.3B 					& 0.98 				& 0.80 				& 0.66 				& 0.84 				& 0.31 				& 0.50 				& 0.68 & -\\
     MUSE-VL~\cite{MUSE-VL} 				& 7B 					& 		-		& 		-		& 		-		& 		-		& 		-		& 		-		& 0.57 & -\\
     Transfusion~\cite{zhou2025transfusion} 					& 3.5B				& - 				& - 				& - 				& - 				& - 				& - 				& 0.63 & -\\
     D-DiT~\cite{dualdiff} 				& 2B 					& 0.97 				& 0.80 				& 0.54 				& 0.76 				& 0.32 				& 0.50 				& 0.65 & -\\
    
            Emu3~\cite{wang2024emu3} & 8B & - & - & - & - & - & - & 0.66 & 110.5 \\
            VILA-U~\cite{vila-u} & 7B & 0.89 & 0.62 & 0.36 & 0.79 & 0.20 & 0.23 & 0.47 & 13.58 \\
            
            \midrule
            Show-o2~\cite{xie2025show} & 1.5B & \textcolor{gray}{0.99} & \textcolor{gray}{0.86} & \textcolor{gray}{0.55} & \textcolor{gray}{0.86} & \textcolor{gray}{0.46} & \textcolor{gray}{0.63} & \textcolor{gray}{0.73} & \textcolor{gray}{10.61}  \\
            \rowcolor{green!10}
            \hspace{0.6em} + \textbf{FlashU (Ours)} & 1.5B & 0.99 & 0.84 & 0.54 & 0.85 & 0.41 & 0.61 & 0.71 & 5.28\textsubscript{+2.01$\times$} \\
            Show-o2~\cite{xie2025show} & 7B & \textcolor{gray}{1.00}  &  \textcolor{gray}{0.87} & \textcolor{gray}{0.58}  & \textcolor{gray}{0.92}  & \textcolor{gray}{0.52}  & \textcolor{gray}{0.62} & \textcolor{gray}{0.76} & \textcolor{gray}{22.74}  \\
            \rowcolor{green!10}
            \hspace{0.6em} + \textbf{FlashU (Ours)} & 7B & 0.99 & 0.80 & 0.53 & 0.89 & 0.46 & 0.58 & 0.72 & 11.82\textsubscript{+1.92$\times$} \\
    
        \bottomrule
    
    \end{tabular}
    }
    \vspace{-1em}
    \end{table*}
\begin{figure*}[!h]
    \centering
    \includegraphics[width=\linewidth]{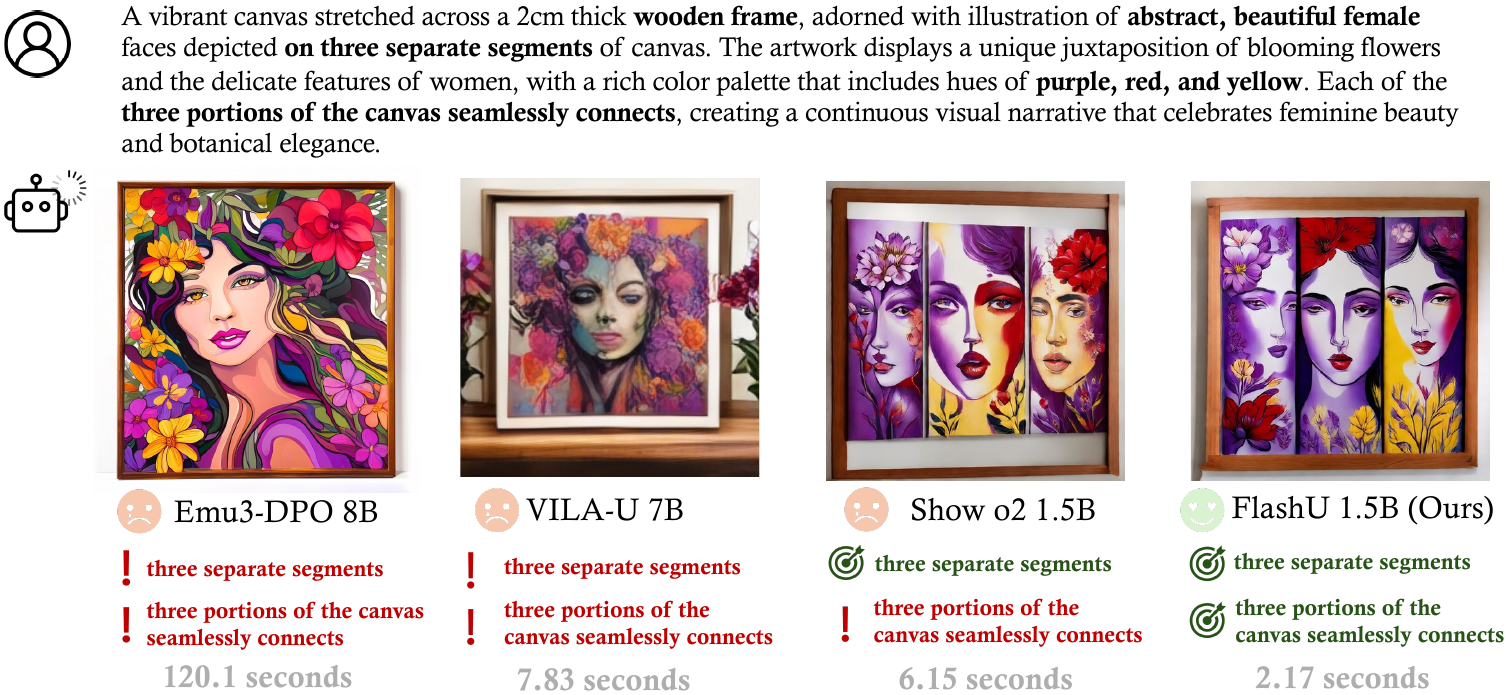}
    \caption{Comparison of four unified multimodal models on generating a three-panel canvas with seamless cross-panel continuity. While baseline models fail to satisfy the structural or continuity requirements, our FlashU 1.5B (from Show-o2~\cite{xie2025show}) accurately produces all three aligned segments with the fastest inference time.}
    \label{fig:case_study_4_models}
    \vspace{-1.0em}
\end{figure*}
\subsection{Experimental Setup}
\label{subsec:exp_setup}

\noindent \textbf{Baselines and Benchmarks.}
To evaluate our method, we compare it against a diverse set of state-of-the-art (SOTA) models. Our primary latency-performance comparison is conducted against native unified models that share architectural similarities with our work, namely VILA-U~\cite{vila-u}, Show-o2~\cite{xie2025show}, and Emu3~\cite{wang2024emu3}. For a more comprehensive assessment, we expand our comparison to include a broader range of advanced systems. This includes other unified models such as MetaMorph~\cite{tong2024metamorph}, ILLUME~\cite{ILLUME}, JanusFlow~\cite{ma2024janusflow}, SynerGen-VL~\cite{synergen-vl}, Liquid~\cite{liquid}, SEED-X~\cite{seed-x}, and Transfusion~\cite{zhou2025transfusion}, as well as leading text-to-image generation models like D-DiT~\cite{dualdiff}, Hunyuan-DiT~\cite{li2024hunyuandit}, Playground v2.5~\cite{Li2024PlaygroundVT}, PixArt-$\Sigma$~\cite{chen2024pixartsigma}, DALL-E 3~\cite{dalle3}, SD3-Medium~\cite{sd3}, TokenFlow-XL$^{*}$~\cite{qu2024tokenflow}, and MUSE-VL~\cite{MUSE-VL}.
For evaluation, we assess multimodal understanding on a suite of benchmarks including MME~\cite{mme}, GQA~\cite{gqa}, MMBench~\cite{mmbench}, MMMU~\cite{mmmu}, MMStar~\cite{mmstar}, and AI2D~\cite{ai2d}. Visual generation capabilities are evaluated using the GenEval~\cite{geneval} and DPG-Bench~\cite{dpgbench}, with detailed results presented in Tables~\ref{tab:geneval_benchmark} and \ref{tab:dpg_benchmark}. For more details on baselines and benchmarks, please refer to Appendix.

\noindent \textbf{Implementation Details.}
Our method is primarily validated on the Show-o2 framework~\cite{xie2025show}. All experiments are conducted using PyTorch on NVIDIA A100-80GB GPUs. Multimodal understanding capabilities are evaluated via the lmms-eval~\citep{zhang2024lmmsevalrealitycheckevaluation}, while generation performance is assessed following the protocols of GenEval and DPG-Bench.

\subsection{Experimental Results} 
\label{subsec:exp_results}

We apply FlashU to the state-of-the-art native unified model, Show-o2~\cite{xie2025show}, to demonstrate its effectiveness across both 1.5B and 7B model scales. As shown in Tables~\ref{tab:mmu_comparison}, \ref{tab:geneval_benchmark}, and \ref{tab:dpg_benchmark}, our method achieves significant latency reduction while maintaining SOTA performance against models with equivalent or larger parameter counts.

\noindent \textbf{Multi-modal Understanding.} Table~\ref{tab:mmu_comparison} presents the impact of FlashU on multimodal understanding benchmarks. The vanilla Show-o2 (7B) model achieves strong SOTA performance but incurs a latency of 1.71 seconds. Our proposed FlashU reduces this to 0.96 seconds, achieving a 1.78$\times$ speedup with minimal performance trade-off. 
Meanwhile, the accelerated FlashU (7B) model maintains its SOTA status against all baselines while demonstrating superior performance-latency trade-offs. 
Specifically, FlashU (7B) outperforms the larger Emu3 (8B) model in both performance (75.3 vs. 58.5 on MMB) and inference speed (0.96s vs. 1.29s). While VILA-U (7B) achieves exceptionally low latency (0.12s), our accelerated model retains substantial performance advantages across all benchmarks (e.g., 45.1 vs. 30.7 on MMMU), validating FlashU as an effective solution for performance-aware deployment.

\begin{table*}[!ht]
    \centering
    \caption{Evaluation on DPG-Bench~\cite{dpgbench}. \# Params. denotes the parameter count of the base LLM. Results in \textcolor{gray}{gray} correspond to vanilla Show-o2. Subscripts indicate the speedup of FlashU. Latency is reported in seconds (s), averaged per sample.}
    \label{tab:dpg_benchmark}
    \setlength\tabcolsep{10pt}
    \renewcommand{\arraystretch}{0.92}
    \resizebox{\linewidth}{!}{%
    \begin{tabular}{lcccccc|cc}
            \toprule
        Method & \# Params. & Global & Entity & Attribute  & Relation  & \textbf{Overall}$\uparrow$ & \textbf{Latency (s)}$\downarrow$ \\
        \midrule
    Hunyuan-DiT~\cite{li2024hunyuandit} 			& 1.5B 			& 84.59 								& 80.59 								& 88.01 								& 74.36 							& 78.87 & -\\
    Playground v2.5~\cite{Li2024PlaygroundVT} 	& - 			& 83.06 								& 82.59 								& 81.20 								& 84.08 							& 75.47 & -\\
    PixArt-$\Sigma$~\cite{chen2024pixartsigma} 	& - 			& 86.89 								& 82.89 								& 88.94 								& 86.59 							& 80.54 & -\\
    DALL-E 3~\cite{dalle3} 						& - 			& 90.97 								& 89.61 								& 88.39 								& 90.58 							& 83.50 & -\\
    SD3-Medium~\cite{sd3} 						& 2B 			& 87.90 								& 91.01 								& 88.83 								& 80.70 							& 84.08 & -\\
    
        Emu3-DPO~\cite{wang2024emu3} & 8B &  - & -  & -  &  - & 81.60 & 124.8 \\
        VILA-U~\cite{vila-u}  & 7B & 89.92 & 82.73 & 80.80 & 87.83 & 74.26 & 9.78 \\
        \midrule
        Show-o2~\cite{xie2025show} & 1.5B & \textcolor{gray}{87.53} & \textcolor{gray}{90.38} & \textcolor{gray}{91.34} & \textcolor{gray}{90.30} & \textcolor{gray}{85.02} & \textcolor{gray}{5.23} \\
        \rowcolor{green!10}
         \hspace{0.6em} + \textbf{FlashU (Ours)} & 1.5B & 84.19 & 89.48 & 90.98 & 90.38 & 84.12 & 2.82\textsubscript{+1.85$\times$} \\
        Show-o2~\cite{xie2025show} & 7B & \textcolor{gray}{89.00} & \textcolor{gray}{91.78} & \textcolor{gray}{89.96} & \textcolor{gray}{91.81} & \textcolor{gray}{86.14} &  \textcolor{gray}{11.30}  \\
        \rowcolor{green!10}
         \hspace{0.6em} + \textbf{FlashU (Ours)} & 7B & 87.79 & 90.30 & 89.29 & 90.30 & 84.39 & 5.89\textsubscript{+1.92$\times$} \\
    
        \bottomrule
    
    \end{tabular}
    }
    \vspace{-1.2em}
    \end{table*}

\noindent \textbf{Image Generation.} We evaluate FlashU on visual generation tasks using the GenEval (Table~\ref{tab:geneval_benchmark}) and DPG-Bench (Table~\ref{tab:dpg_benchmark}). Both benchmarks consistently demonstrate that FlashU provides substantial acceleration (approximately 1.9$\times$ to 2.0$\times$) while preserving the original model's SOTA capabilities. On GenEval, FlashU achieves a 1.92$\times$ speedup for the 7B model (22.74s to 11.82s) and a 2.01$\times$ speedup for the 1.5B model (10.61s to 5.28s). The accelerated 7B model's Overall score (0.72) remains superior to other unified baselines, including Emu3 (0.66) and VILA-U (0.47). On DPG-Bench, FlashU delivers a 1.92$\times$ speedup for the 7B model (11.30s to 5.89s) and a 1.85$\times$ speedup for the 1.5B model (5.23s to 2.82s). Despite this significant latency reduction, the accelerated 7B model's Overall score of 84.39 remains highly competitive, surpassing SOTA generation-specific models like SD3-Medium (84.08) and decisively outperforming VILA-U (74.26).

\noindent \textbf{Efficiency and Performance Summary.} Across all tasks and model scales, FlashU consistently delivers inference acceleration ranging from 1.78$\times$ to 2.01$\times$ without compromising the original Show-o2 model's SOTA status. Our accelerated framework maintains a clear performance lead over competing unified models while operating at a fraction of the computational cost, demonstrating an effective and practical solution for deploying large-scale unified models.
Meanwhile, as shown in Figure~\ref{fig:case_study_4_models}, our accelerated framework produces generations that remain highly faithful to the prompt, even surpassing the vanilla model in instruction following and matching every specified detail.

\subsection{Ablation Study}
\label{subsec:ablation}
To validate the individual contributions of our components, we conducted an ablation study on both the 7B and 1.5B models using the DPG-Bench~\cite{dpgbench} benchmark. Results are detailed in Tables~\ref{tab:7b_results_corrected} and \ref{tab:1.5b_results_corrected} for the 7B and 1.5B models, respectively. We establish FFN pruning as the acceleration baseline, achieving an overall speedup of around 1.3$\times$. Building upon this static pruning, our dynamic technologies account for the majority of the acceleration: Adaptive Guidance and Dynamic Layer Pruning jointly yield substantial additional speedup. The study also highlights the critical role of our Hybrid Network module. Disabling this module results in a significant score drop, demonstrating its essential quality-preserving function in restoring fidelity.

\begin{table}[!h]
\centering
\caption{Ablation study on the 7B model of Show-o2 using DPG-Bench~\cite{dpgbench}. The subscript values show the speedup relative to Show-o2 baseline. Latency is measured in seconds (s).}
\label{tab:7b_results_corrected}
\begin{tabularx}{\columnwidth}{lcc}
    \toprule
    \textbf{Strategy} & \textbf{Latency (s)} & \textbf{Score}\\
    \midrule
    Show-o2~\cite{xie2025show} & 11.30 & 86.14  \\
    \midrule
    FlashU (Ours) & 5.89\textsubscript{+1.92$\times$} & 84.39 \\
    \quad \textit{w/o Dynamic Layer Pruning} & 6.70\textsubscript{+1.69$\times$} & 85.56\\
    \quad \textit{w/o Hybrid Network} & 5.65\textsubscript{+2.00$\times$} & 83.68  \\
    \quad \textit{w/o Adaptive Guidance} & 8.86\textsubscript{+1.28$\times$} & 85.34\\
    \quad \textit{FFN Pruning Only} & 9.21\textsubscript{+1.23$\times$} & 85.08 \\
    \bottomrule
\end{tabularx}
\end{table}

\begin{table}[!ht]
\centering
\caption{Ablation study on the 1.5B model of Show-o2 using DPG-Bench~\cite{dpgbench}. The subscript values show the speedup relative to Show-o2 baseline. Latency is measured in seconds (s).}
\label{tab:1.5b_results_corrected}
\begin{tabularx}{\columnwidth}{lcc}
    \toprule
    \textbf{Strategy} & \textbf{Latency (s)} & \textbf{Score}\\
    \midrule
    Show-o2~\cite{xie2025show} & 5.23 & 85.02 \\
    \midrule
    FlashU (Ours) & 2.80\textsubscript{+1.87$\times$} & 84.12\\
    \quad \textit{w/o Dynamic Layer Pruning} & 2.91\textsubscript{+1.80$\times$} & 85.33 \\
    \quad \textit{w/o Hybrid Network} & 2.48\textsubscript{+2.11$\times$} & 83.57 \\
    \quad \textit{w/o Adaptive Guidance} & 3.92\textsubscript{+1.33$\times$} & 83.42\\
    \quad \textit{FFN Pruning Only} & 3.84\textsubscript{+1.36$\times$} & 84.27 \\
    \bottomrule
\end{tabularx}
\vspace{-1em}
\end{table}



\section{Conclusion}
\label{sec:conclusion}


We present a fundamental shift in accelerating unified multimodal models through systematic analysis revealing that these models, despite their monolithic architecture, inherently develop task-specific computational pathways. This discovery challenges conventional uniform acceleration strategies and establishes task-aware optimization that aligns with the model's internal mechanisms. Our proposed \textbf{FlashU} demonstrates that respecting this natural specialization enables 1.78$\times$ to 2.01$\times$ speedup while maintaining SOTA performance across understanding and generation tasks. These results validate that unified models benefit from task-specific acceleration rather than one-size-fits-all approaches. As unified models continue to scale, task-aware acceleration grounded in deep understanding of model internals represents a promising direction for practical real-world deployment. Moreover, our analysis provides a unified lens for interpreting efficiency behaviors observed across diverse multimodal architectures. FlashU further highlights that efficient inference does not require sacrificing representational richness when model structure is properly leveraged. We believe this paradigm will inspire future research on principled, model-aligned acceleration methods for next generation of unified multimodal systems.
\section*{Acknowledgements}
This work was supported by the Shanghai Science and Technology Program (Grant No. 25ZR1402278) and WUYING - Alibaba Cloud.

{
    \small
    \bibliographystyle{ieeenat_fullname}
    \bibliography{main}
}


\appendix
\maketitlesupplementary
\setcounter{page}{1}

\begin{table*}[!ht]
    \centering
    \caption{Latency Comparison on DPG-Bench~\cite{dpgbench}. Results in \textcolor{gray}{gray} correspond to vanilla Show-o2. Subscripts indicate the speedup of FlashU. Latency is reported in seconds (s), averaged per sample.}
    \label{tab:dpg_benchmark_latency}
    \setlength\tabcolsep{10pt}     
    \renewcommand{\arraystretch}{0.92} 
    \resizebox{\linewidth}{!}{%
    \begin{tabular}{lcccccc|c}
        \toprule

        Method & \# Params. & Global & Entity & Attribute  & Relation  & \textbf{Overall}$\uparrow$ & \textbf{Latency (s)}$\downarrow$ \\
        \midrule
        Hunyuan-DiT~\cite{li2024hunyuandit} 			& 1.5B 			& 84.59 								& 80.59 								& 88.01 								& 74.36 							& 78.87 & 20.1 \\
        Playground v2.5~\cite{Li2024PlaygroundVT} 	& - 			& 83.06 								& 82.59 								& 81.20 								& 84.08 							& 75.47 & 7.67\\
        TokenFlow-t2i~\cite{qu2024tokenflow} & 7B & 72.64 & 74.72 & 77.67 & 83.80 & 67.44 & 0.97 \\
         JanusFlow~\cite{ma2024janusflow} & 1.5B & 81.16 & 85.16 & 85.32 & 91.05 & 77.57 & 1.09 \\
        VILA-U~\cite{vila-u}  & 7B & 89.92 & 82.73 & 80.80 & 87.83 & 74.26 & 9.78 \\
        Liquid~\cite{liquid} & 8B   & 79.64 & 87.61 & 85.24 & 91.01 & 80.26 & 37.05 \\
        Emu3-DPO~\cite{wang2024emu3} & 8B &  - & -  & -  &  - & 81.60 & 124.8 \\

        \midrule
        Show-o2~\cite{xie2025show} & 1.5B & \textcolor{gray}{87.53} & \textcolor{gray}{90.38} & \textcolor{gray}{91.34} & \textcolor{gray}{90.30} & \textcolor{gray}{85.02} & \textcolor{gray}{5.23} \\
        \rowcolor{green!10}
         \hspace{0.6em} + \textbf{FlashU (Ours)} & 1.5B & 84.19 & 89.48 & 90.98 & 90.38 & 84.12 & 2.82\textsubscript{+1.85$\times$} \\
        Show-o2~\cite{xie2025show} & 7B & \textcolor{gray}{89.00} & \textcolor{gray}{91.78} & \textcolor{gray}{89.96} & \textcolor{gray}{91.81} & \textcolor{gray}{86.14} &  \textcolor{gray}{11.30}  \\
        \rowcolor{green!10}
         \hspace{0.6em} + \textbf{FlashU (Ours)} & 7B & 87.79 & 90.30 & 89.29 & 90.30 & 84.39 & 5.89\textsubscript{+1.92$\times$} \\
        \bottomrule
    \end{tabular}
    }





\end{table*}

\section{Experimental Setup}
\label{sec:setup}

We evaluate FlashU against state-of-the-art (SOTA) unified and specialized models across multimodal understanding and generation tasks.

\subsection{Baselines}

Following the taxonomy proposed in Show-o2~\cite{xie2025show}, we classify baselines into three categories:

\begin{itemize}
    \item \textbf{Native Unified Models:} Architectures that integrate understanding and generation within a single framework. We compare against \textit{Show-o2}~\cite{xie2025show}, \textit{VILA-U}~\cite{vila-u}, \textit{Emu3}~\cite{wang2024emu3}, \textit{Transfusion}~\cite{zhou2025transfusion}, \textit{SynerGen-VL}~\cite{synergen-vl}, \textit{Liquid}~\cite{liquid}, \textit{D-DiT}~\cite{dualdiff}, and \textit{MUSE-VL}~\cite{MUSE-VL}.
    \item \textbf{Assembling Tailored Models:} Systems that achieve unification by pipelining specialized experts (e.g., separate LLMs and diffusion models). Baselines include \textit{MetaMorph}~\cite{tong2024metamorph}, \textit{ILLUME}~\cite{ILLUME}, \textit{JanusFlow}~\cite{ma2024janusflow}, \textit{SEED-X}~\cite{seed-x}, and \textit{TokenFlow-XL}~\cite{qu2024tokenflow}.
    \item \textbf{Specialized Generation Models:} To benchmark generation fidelity against domain experts, we include \textit{Hunyuan-DiT}~\cite{li2024hunyuandit}, \textit{PixArt-$\Sigma$}~\cite{chen2024pixartsigma}, \textit{SD3-Medium}~\cite{sd3}, \textit{Playground v2.5}~\cite{Li2024PlaygroundVT}, and \textit{DALL-E 3}~\cite{dalle3}.
\end{itemize}

Our primary latency-performance comparison is conducted against open-source native unified models that share architectural similarities with our work, namely VILA-U~\cite{vila-u}, Show-o2~\cite{xie2025show}, and Emu3~\cite{wang2024emu3}. To facilitate a more comprehensive analysis, we present an extended latency-performance comparison with \textit{Liquid}~\cite{liquid}, \textit{TokenFlow-XL}~\cite{qu2024tokenflow}, and \textit{JanusFlow}~\cite{ma2024janusflow} in Table~\ref{tab:dpg_benchmark_latency} and Figure~\ref{fig:latency_performance_tradeoff} of the Appendix. 

\begin{figure}[t]
    \centering
    \includegraphics[width=\linewidth, trim=1.75cm 0.1cm 2.9cm 0.2cm, clip]{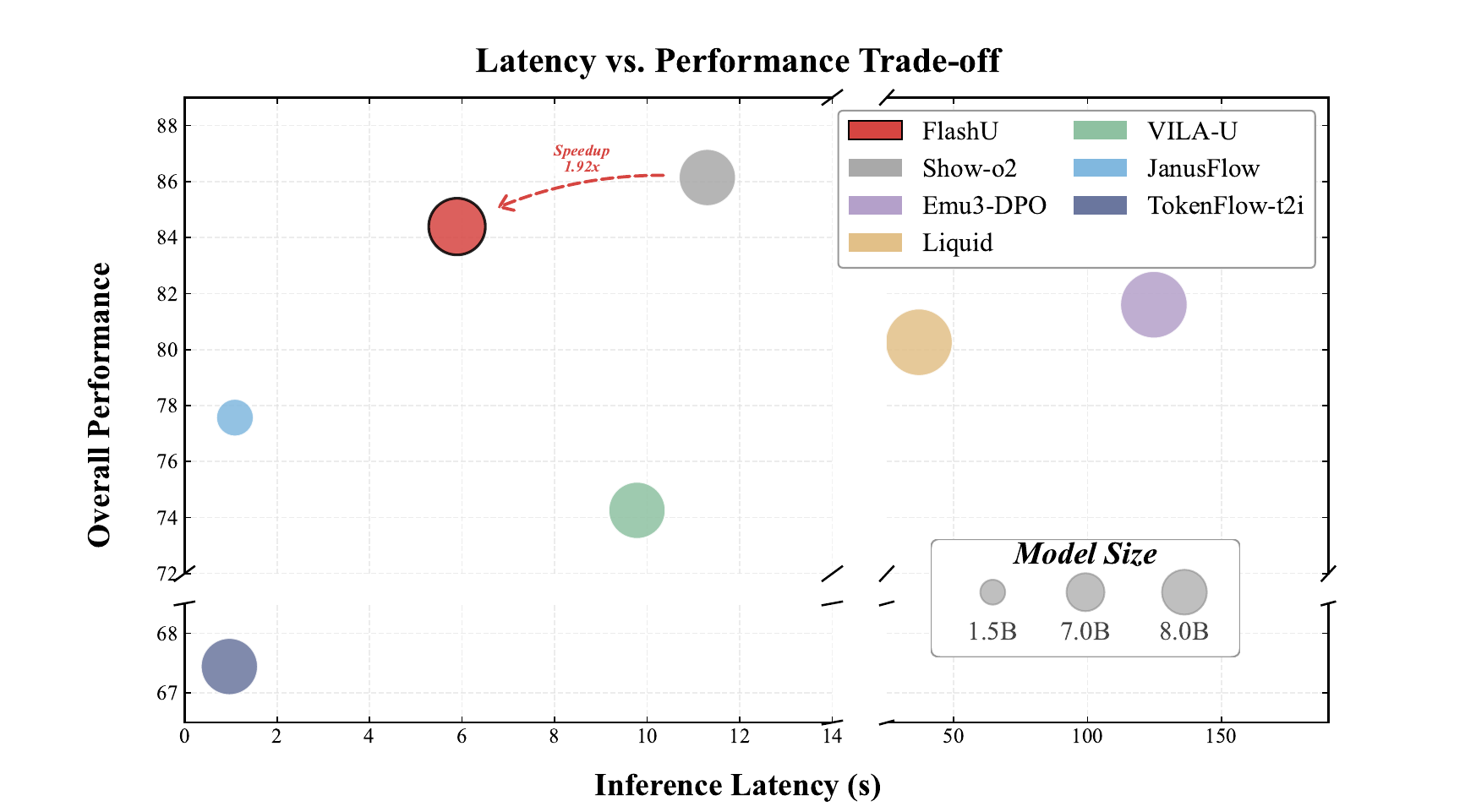}
    \caption{\textbf{Latency vs. Performance Trade-off.} 
    Comparison of FlashU against native unified models (Show-o2, VILA-U, Emu3-DPO, Liquid) and tailored models (JanusFlow, TokenFlow-t2i). 
    The x-axis represents inference latency (lower is better), and the y-axis represents overall performance on DPG-Bench(higher is better). 
    Circle sizes denote model parameter counts (1.5B, 7B, 8B). 
    }
    \vspace{-4mm}
    \label{fig:latency_performance_tradeoff}
\end{figure}

\subsection{Benchmarks and Datasets}

\paragraph{Multimodal Understanding} We utilize the \texttt{lmms-eval} suite to assess performance across six standard benchmarks:

\begin{itemize}
    \item \textbf{MME}~\cite{mme}: Evaluates perception and cognition using a strict Yes/No format to decouple reasoning capabilities from instruction-following bias.
    
    \item \textbf{GQA}~\cite{gqa}: Tests compositional reasoning via scene graph-based queries, requiring the parsing of multi-hop spatial and semantic relationships.
    
    \item \textbf{MMBench}~\cite{mmbench}: Addresses evaluation robustness. It employs a ``Circular Evaluation'' strategy (shuffling options) to penalize inconsistency and random guessing.
    
    \item \textbf{MMMU}~\cite{mmmu}: Assesses expert-level reasoning across diverse disciplines, utilizing college-level problems that demand subject-matter expertise.
    
    \item \textbf{MMStar}~\cite{mmstar}: Focuses on visual dependency by filtering out samples answerable via text priors or world knowledge alone (``blind answering'').
    
    \item \textbf{AI2D}~\cite{ai2d}: Examines diagrammatic reasoning, requiring the interpretation of synthetic layouts, symbolic pointers, and process flows in scientific textbook illustrations.
\end{itemize}

\paragraph{Visual Generation} We evaluate alignment and fidelity using two specialized benchmarks designed to test prompt adherence beyond standard metrics:

\begin{itemize}
    \item \textbf{GenEval}~\cite{geneval}: An object-focused evaluation framework that treats generation assessment as a detection problem. It utilizes a fixed set of prompts and pre-trained object detectors to verify specific compositional properties, including \textit{Single/Two Object} presence, \textit{Counting} precision, \textit{Color} attribute binding, and \textit{Spatial Position}.
    
    \item \textbf{DPG-Bench}~\cite{dpgbench}: Introduced in ELLA~\cite{dpgbench}, this benchmark employs over 1000 dense, highly descriptive prompts to assess the model's ability to handle high information loads. It evaluates fine-grained alignment across four specific dimensions: \textit{Global} scene consistency, \textit{Entity} presence, \textit{Attribute} accuracy, and \textit{Relation} (interaction between objects).
\end{itemize}

\section{Implementation Details}
\label{sec:implementation}

In this section, we provide a comprehensive overview of the FlashU inference protocol. We first visually detail the four core acceleration method introduced in our framework, followed by the complete pseudocode governing the dynamic dispatching of these mechanisms and the specific hyperparameter configurations.

\subsection{Pseudocode for Unified FlashU Inference}

We present the unified inference protocol of FlashU in Algorithm \ref{alg:unified}. This algorithm demonstrates how the framework dynamically dispatches input queries to task-specific acceleration pathways—switching between iterative generation optimization and understanding optimization within a single unified architecture.

\subsection{Hyperparameter Settings}

In this section, we provide the specific hyperparameter values used for Generation and Multimodal Understanding (MMU) tasks in Table \ref{tab:hyperparams_gen} and Table \ref{tab:hyperparams_mmu}, respectively. Unless otherwise stated, all notation aligns with the definitions in the main text. Baseline and limit constants are defined as follows: $T_0$ denotes the total inference steps in the baseline Show-o2 model; $s_0$ represents the base guidance scale; and $N_{max}$ indicates the maximum number of generated tokens for Question Answering (QA) tasks.

\begin{table}[h]
    \centering
    \caption{Hyperparameters for Generation Tasks.}
    \label{tab:hyperparams_gen}
    \begin{tabular}{lcc}
        \toprule
        \textbf{Hyperparameter} & \textbf{Symbol} & \textbf{Value} \\
        \midrule
        Total Inference Steps & $T$ & $0.7 T_0$ \\
        Hybrid Threshold & $\tau$ & 0.3\\
        FFN Pruning Ratio & $r_p$ & 0.20 \\
        Layer Skipping Ratio & $r_{LS}$ & 0.20 \\
        Recalculation Interval & $T_{LS}$ & 10 \\
        Diffusion Head Cache Interval & $\mathcal{T}_{cache}$ & 5 \\
        Switching Timestep & $t_{switch}$ & $0.3 T$ \\
        Low Guidance Scale & $s_{low}$ & $0.75 s_0$ \\
        High Guidance Scale & $s_{high}$ & $s_0$ \\
        \bottomrule
    \end{tabular}
\end{table}

\begin{table}[h]
    \centering
    \caption{Hyperparameters for Multimodal Understanding Tasks.}
    \label{tab:hyperparams_mmu}
    \begin{tabular}{lcc}
        \toprule
        \textbf{Hyperparameter} & \textbf{Symbol} & \textbf{Value} \\
        \midrule
        Max Generated Tokens & $N_{max}$ & 128 \\
        Hybrid Threshold & $\tau$ & 0.3\\
        FFN Pruning Ratio & $r_p$ & 0.10 \\
        Layer Skipping Ratio & $r_{LS}$ & 0.10 \\
        Calibration Samples & $N_{calib}$ & 20 \\
        Recalculation Interval & $T_{LS}$ & 10 \\
        Shallow layer to prune & $L_{prune}$ & 2\\
        Token Pruning Ratio & \(\rho\) & 0.50 \\
        \bottomrule
    \end{tabular}
\end{table}
\section{Parameter Sensitivity Analysis}
\label{sec:sensitivity_analysis}

In this section, we conduct a comprehensive sensitivity analysis to evaluate the impact of key hyperparameters on both inference latency and generation quality. Specifically, we investigate the FFN Pruning Ratio ($r_p$), Layer Skipping Ratio ($r_{LS}$), and the Hybrid Threshold ($\tau$) under different guidance strategies. We performed this analysis on the 1.5B model of Show-o2, reporting the DPG-Bench score and predicted speedup relative to the baseline. 

\subsection{Impact of FFN Pruning Ratio}
\label{subsec:ffn_pruning}
\begin{figure}[h!]
    \centering
    \vspace{-5mm}

    \includegraphics[width=0.8\linewidth]{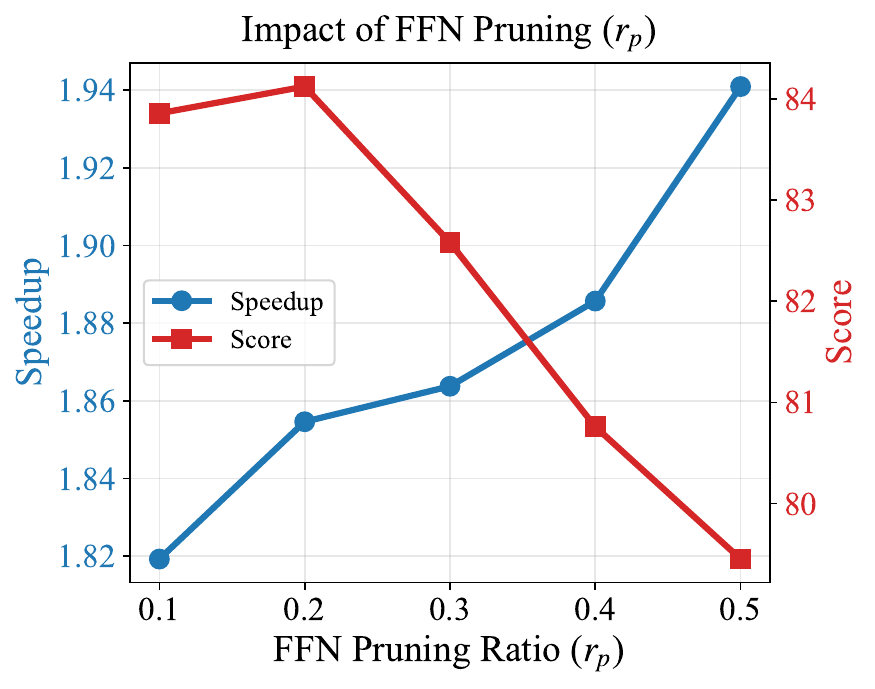}
    \vspace{-2.5mm}

    \caption{\textbf{Sensitivity Analysis of FFN Pruning Ratio.} The trade-off between generation quality (DPG-Bench score) and inference speedup is shown as the FFN pruning ratio ($r_p$) varies from 0.1 to 0.5.}
    \label{fig:sensitivity_ffn}
\end{figure}
As shown in Figure~\ref{fig:sensitivity_ffn}, we examine the model's sensitivity to the FFN Pruning Ratio ($r_p$) under a fixed configuration of $\tau=0.3$ and $r_{LS}=0.2$, utilizing the adaptive guidance strategy.

As $r_p$ increases from $0.1$ to $0.5$, the speedup exhibits a moderate monotonic increase, rising from $1.82\times$ to $1.94\times$. However, the generation quality, measured by the DPG-Bench score, reveals a distinct non-linear trend. The score peaks at $84.12$ when $r_p=0.2$, defining the optimal performance point. Beyond this point, increasing $r_p$ to $0.4$ and $0.5$ leads to a noticeable degradation in quality, with the score dropping to $79.45$ at $r_p=0.5$. Further aggressive pruning yields diminishing returns in acceleration while significantly compromising visual fidelity.

\subsection{Impact of Layer Skipping Ratio}
\label{subsec:layer_skipping}
\begin{figure}[h!]
    \centering
    \vspace{-5mm}

    \includegraphics[width=0.8\linewidth]{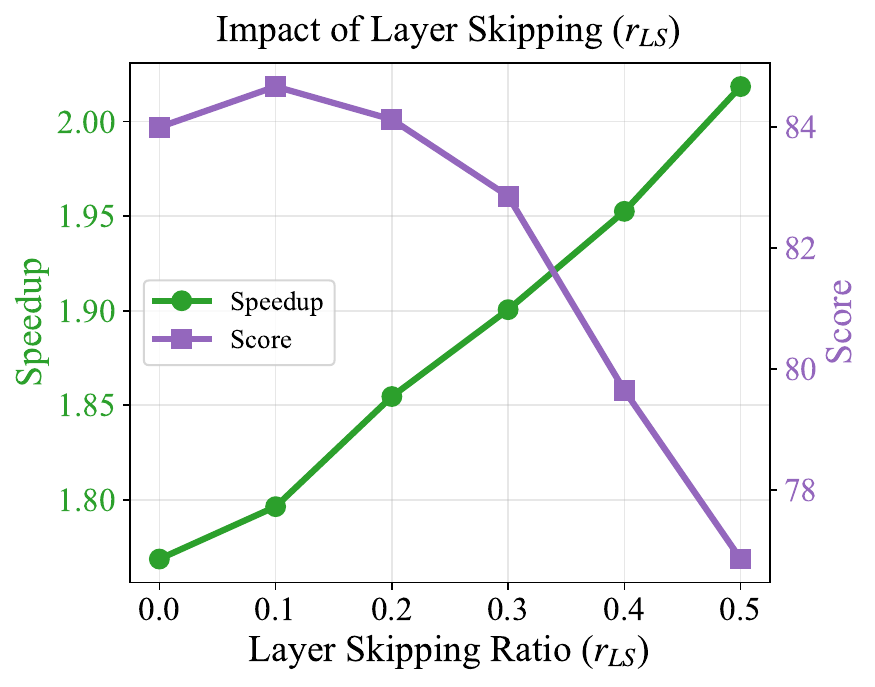}
    \vspace{-2.5mm}

    \caption{\textbf{Sensitivity Analysis of Layer Skipping Ratio.} The relationship between generation quality and speedup across different layer skipping ratios ($r_{LS}$).}
    \label{fig:sensitivity_layer}
\end{figure}
We further analyze the effect of $r_{LS}$ by varying it from $0$ to $0.5$, while maintaining $r_p=0.2$ and $\tau=0.3$.

As shown in Figure~\ref{fig:sensitivity_layer}, the speedup demonstrates a strong positive correlation with $r_{LS}$, improving significantly from $1.77\times$ achieved at $r_{LS}=0$ to $2.02\times$ at $r_{LS}=0.5$. In terms of quality, the model exhibits robustness when $r_{LS} \le 0.2$, with the DPG-Bench score remaining stable and peaking at $84.66$ when $r_{LS}=0.1$. However, a sharp performance collapse is observed when $r_{LS}$ exceeds $0.3$; specifically, at $r_{LS}=0.5$, the score plummets to $76.86$.

Setting $r_{LS}$ to $0.2$ establishes the upper bound for layer skipping. Although this setting incurs a negligible quality trade-off compared to the peak at $r_{LS}=0.1$, it guarantees a substantial $1.85\times$ acceleration while maintaining robust generative capability.

\subsection{Trade-off between Hybrid Threshold and Adaptive Guidance Strategy}
\label{subsec:hybrid_threshold}
\begin{figure}[h!]
\vspace{-5mm}
    \centering
    \includegraphics[width=\linewidth]{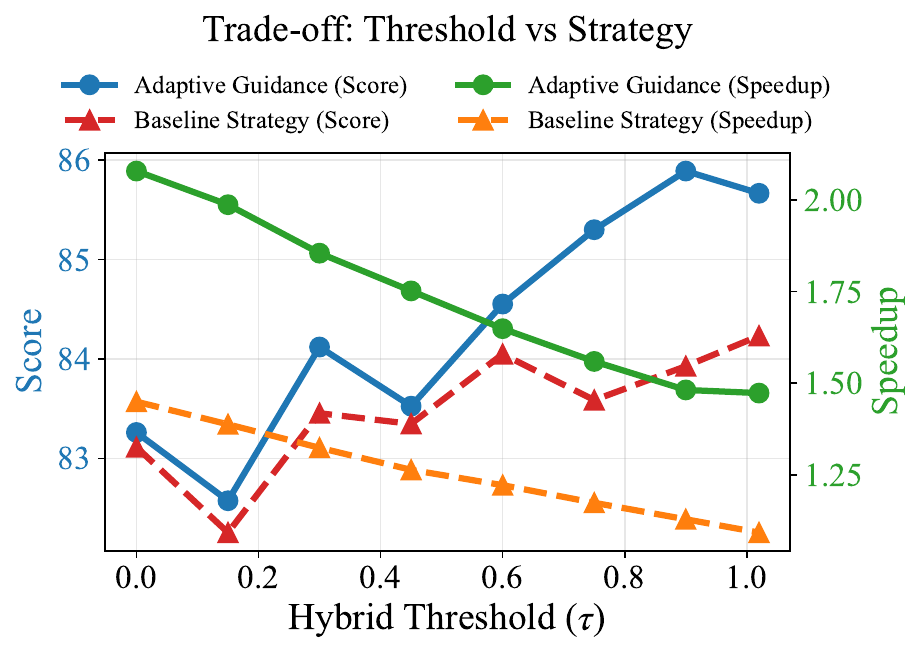}
\vspace{-2.5mm}

    \caption{\textbf{Comparative Analysis of Hybrid Threshold Strategies.} The efficiency-quality trade-off curves for Adaptive Guidance versus Baseline strategies across different hybrid threshold ($\tau$) values. }
    \label{fig:sensitivity_hybrid}

\end{figure}
We also investigate the trade-off between efficiency and quality by varying the Hybrid Threshold ($\tau$) across two strategies: the proposed Adaptive Guidance with reduced steps and the Baseline as shown in Figure~\ref{fig:sensitivity_hybrid}. 

Both strategies follow a consistent trade-off pattern: as $\tau$ increases (retaining more computation), the speedup decreases while the generation quality improves.

The Adaptive Guidance demonstrates a superior Pareto frontier compared to the Baseline. At comparable quality levels, Adaptive Guidance consistently achieves higher speedups. For instance, at $\tau \approx 0.3$, Adaptive Guidance achieves $1.85\times$ speedup versus $1.32\times$ for the Baseline, achieving a score of $84.12$ compared to the Baseline's $83.45$. Adaptive Guidance not only accelerates inference but also achieves a higher peak quality. At higher thresholds, such as $\tau \approx 0.9$, Adaptive Guidance reaches a score of $85.89$, significantly outperforming the Baseline's peak performance of roughly $84.23$. The combination of Adaptive Guidance and reduced sampling steps proves to be a robust strategy, effectively expanding the efficiency-quality operating envelope beyond the standard baseline.

Based on the analysis above, we adopt the configuration of $r_p=0.2$, $r_{LS}=0.2$, and $\tau=0.3$ with Adaptive Guidance as our default setting. This combination strikes an optimal balance, delivering a $1.85\times$ speedup while maintaining a competitive DPG-Bench score of $84.12$.



\begin{algorithm}[h!]
\caption{Unified FlashU Inference Protocol}
\label{alg:unified}
\begin{algorithmic}[1]

\Input Task Indicator $\mathcal{T} \in \{\textsc{Gen}, \textsc{Und}\}$, Input Query $\mathcal{Q}$ (Text/Image), Unified Model Weights $\theta$.
\Output Generated Response $\mathcal{R}$ (Image or Text).

\Initialize 
\State Apply Task-Specific Network Pruning. 
\State $\mathcal{L}_{skip} \leftarrow \emptyset$; $H_{cache} \leftarrow \text{None}$; \State $\mathcal{R} \leftarrow \emptyset$.

\If{$\mathcal{T} == \textsc{Gen}$} \Comment{\textcolor{BrickRed}{\textbf{Pathway A: Image Generation}}}
    \State \textbf{Parse} $\mathcal{Q}$ to initial noise $z_T$ and text condition $c$.
      
    \For{$t = T$ \textbf{to} $1$} \Comment{Iterative Denoising Process}
        
        \CommentLine{\textcolor{Violet}{Adaptive Guidance}}

       \State $s_t \leftarrow (t > t_{switch}) ? s_{low} : s_{high}$
        \State $RecalcFlag \leftarrow (t \pmod{K_{skip}} == 0)$
       
        \State $h \leftarrow \text{Embed}(z_t, c)$
        \For{$l = 1$ \textbf{to} $L$}
            \If{$l \in \mathcal{L}_{skip}$ \textbf{and not} $RecalcFlag$}
                \State \textbf{continue}
            \EndIf
           
            \State $h_{in} \leftarrow h$; $h \leftarrow \text{SelfAttention}(h, \text{Mask}_{gen})$
\State $h  \leftarrow (t > \tau T) ?  h + \text{FFN}_{p}(h) : h + \text{FFN}_{f}(h)$
            \If{$RecalcFlag$ \textbf{and} $S(h_{in}, h) > \delta_{th}$}
                \State $\mathcal{L}_{skip}.\text{add}(l)$
            \EndIf
        \EndFor

        \CommentLine{\textcolor{Violet}{Diffusion Head Cache}}
        \If{$t \pmod{K_{cache}} == 0$}
            \State $H_{cache} \leftarrow \text{DiffHead}(h, t)$
        \EndIf
        \State $z_{t-1} \leftarrow \text{SchedulerStep}(z_t, H_{cache}, s_t)$
    \EndFor
    \State $\mathcal{R} \leftarrow \text{VAE-Decode}(z_0)$
\ElsIf{$\mathcal{T} == \textsc{Und}$} \Comment{\textcolor{RoyalBlue}{\textbf{Pathway B: MMU}}}
    \State \textbf{Parse} $\mathcal{Q}$ to visual tokens $X_v$ and text tokens $X_t$.

    \For{$t = 1$ \textbf{to} $N_{max}$} \Comment{Autoregressive Generation}
        \State $H \leftarrow \text{Embed}(X_v, X_t, \mathcal{R})$ 
        
        \For{$l = 1$ \textbf{to} $L$} 
            \If{$l \in \mathcal{L}_{skip}$ \textbf{and not} $RecalcFlag$}
                \State \textbf{continue}
            \EndIf

            \CommentLine{\textcolor{Violet}{Dynamic Visual Token Pruning}}
            \If{$l == L_{prune}$}
                \State $Q, K, V \leftarrow \text{Project}(H)$
                \State $S_{imp} \leftarrow \|V\|_2$ \Comment{\textcolor{Violet}{V-Norm proxy}}
                \State $\mathcal{I}_{keep} \leftarrow \text{TopK}(S_{imp}, (1-\rho)|X_v|)$

                \State $H \leftarrow \text{Gather}(\mathcal{I}_{keep}\cup{X_v}, X_t, \mathcal{R})$

            \EndIf

            \State $H \leftarrow \text{TransformerBlock}(H)$
        \EndFor
        
        \State $w_{n} \leftarrow \text{Sample}(\text{LM-Head}(H[-1]))$ 

        \State \textbf{break} \textbf{if} $w_{n} == \textsc{EOS}$ \Comment{Stopping Criterion}
        \State $\mathcal{R}.\text{append}(w_{next})$ \Comment{Update context}
    \EndFor
\EndIf

\State \Return $\mathcal{R}$
\end{algorithmic}
\end{algorithm}

\section{Component-wise Acceleration Analysis}
\label{subsec:component_analysis}

\begin{figure}[!ht]
    \centering
    \includegraphics[width=0.9\linewidth]{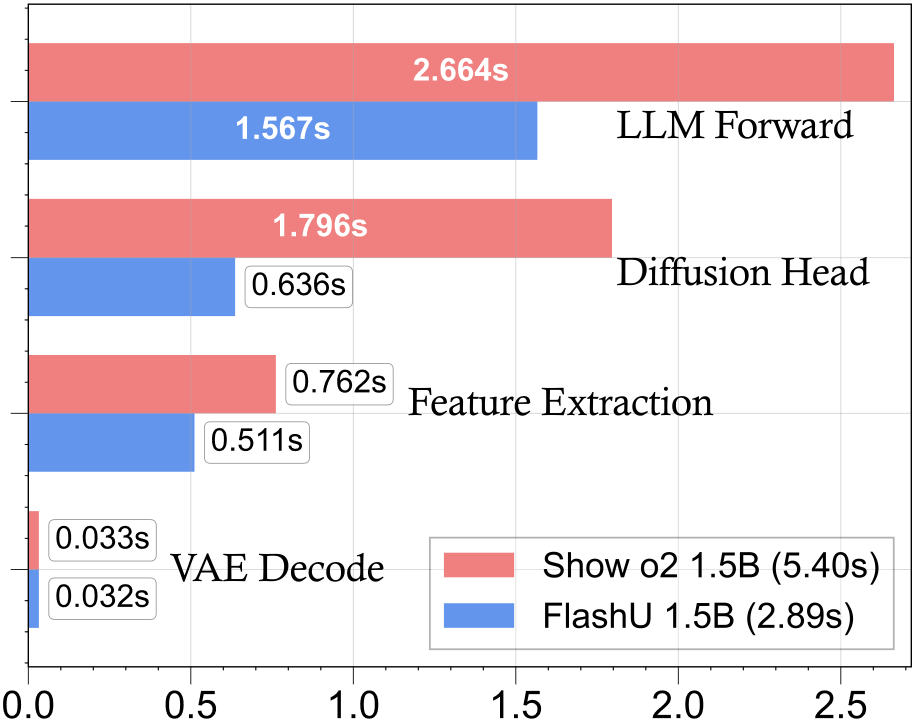}

    \caption{\textbf{Component-wise Latency Profiling Analysis.} Inference time breakdown by component for Show o2 and FlashU on the 1.5B model.}
    \label{fig:time_profiling_analysis_1_5B}
    \vspace{-1.0em}
\end{figure}

To understand where FlashU's acceleration comes from, we conduct fine-grained profiling of the inference time breakdown by component on the 1.5B model. As shown in Figure~\ref{fig:time_profiling_analysis_1_5B}, the analysis reveals heterogeneous acceleration across components: the Diffusion Head achieves the highest relative speedup (2.82$\times$), primarily from iterative caching and adaptive guidance. The LLM forward pass, being the largest component (49\% of total time), achieves 1.70$\times$ speedup through FFN pruning and dynamic layer skipping, contributing the most absolute time savings. Feature Extraction shows moderate acceleration (1.49$\times$) via the Hybrid FFN's lightweight path. VAE decoding remains unchanged, confirming our optimizations introduce no overhead. This component-wise analysis validates that our multi-component optimization strategy effectively eliminates bottlenecks across the entire inference pipeline.

\section{Additional Qualitative Results}
\label{sec:appendix_qualitative}

\begin{figure*}[htbp]
    \centering

\centering

    \includegraphics[width=0.9\linewidth, trim=0cm 1.85cm 0cm 0.1cm, clip]{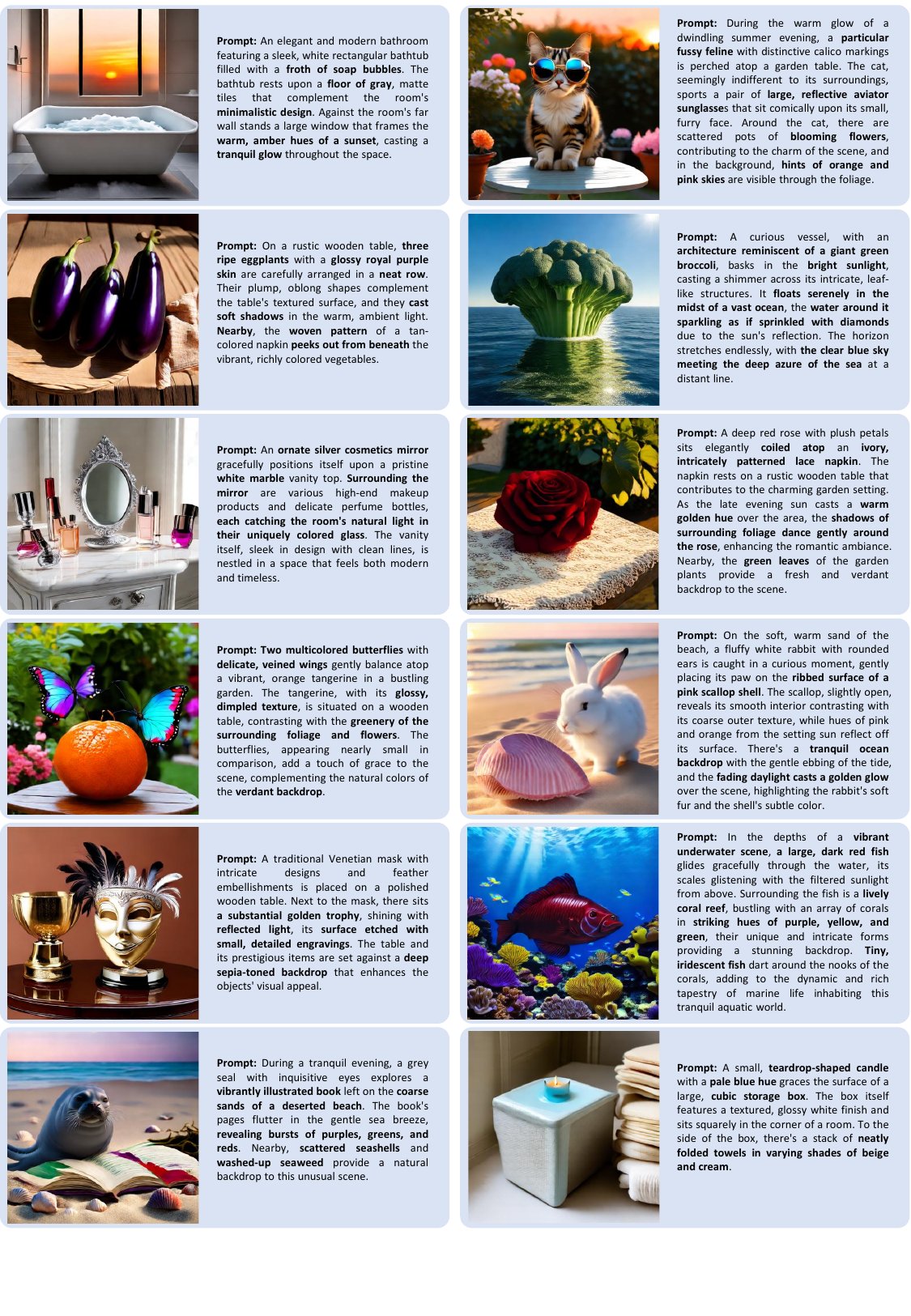}
    \caption{
        \textbf{Qualitative results of text-to-image generation.} 
        We showcase samples generated by FlashU across diverse prompts. 
        The model successfully handles complex descriptions, rendering high-quality textures and correct object placements, demonstrating that our acceleration framework preserves the generative performance of the backbone model.
    }
    \label{fig:visualization}
\end{figure*}

We present additional qualitative results in Figure~\ref{fig:visualization} to further demonstrate the generation capabilities of FlashU. As shown, our method generates high-fidelity and photorealistic images that accurately align with the given text prompts, effectively preserving the visual quality of the baseline model while significantly accelerating the inference process.

\end{document}